\documentclass[runningheads]{llncs}
\usepackage[T1]{fontenc}
\usepackage{graphicx}
\usepackage{graphicx,booktabs,bbding,pifont,amsmath,multirow,CJKutf8,hyperref,amssymb,flushend}

\begin{document}

\pagestyle{plain}
\title{DDVT: Dynamic Dual-level Vision Transformer Fusion Network for Answer Grounding in Visual Question Answering}
%
%

\author{
\makebox[\textwidth][l]{%
Yue Zhang\textsuperscript{1} \and
Xiangyu Li\textsuperscript{1} \and
Wanshu Fan\textsuperscript{1$\ast$} \and
Xin Yang\textsuperscript{2} \and
Dongsheng Zhou\textsuperscript{1,2$\ast$}
}}

\authorrunning{Yue Zhang et al.}
%
\institute{
    \textsuperscript{}National and Local Joint Engineering Laboratory of Computer Aided Design, School of Software Engineering, Dalian University, Dalian, China \\
    \and
    \textsuperscript{}School of Computer Science and Technology, Dalian University of Technology, Dalian, China \\
}
\begingroup
\renewcommand{\thefootnote}{}
\footnotetext{%
\makebox[0pt][l]{%
\hspace*{-1.2em}
\parbox{\linewidth}{%
\noindent *Corresponding author\\
\noindent Email address: fanwanshu@dlu.edu.cn (Wanshu Fan);
zhouds@dlu.edu.cn (Dongsheng Zhou)
}}}
\endgroup

\maketitle              
\begin{abstract}
Answer grounding in visual question answering aims to locate the region from a given natural language question associated with the visual content of an image, which has garnered significant attention due to its practical applications. 
In this paper, we introduce the Dynamic Dual-level Vision Transformer Fusion Network (DDVT) for answer grounding in visual question answering.
Specifically, we propose a question-guided dynamic regional-level module (QGDR) that combines complementary image context through ROI Align and text content, enabling precise localization of text-related visual content. 
Moreover, we present a cross-modal multi-scale aggregation module (CMA) that enhances feature fusion between pixel-level and region-level features, facilitating the effective localization of visual content associated with grounded answers. 
Furthermore, we fuse the located visual content with text features to locate the region and provide answers to questions posed about the image.
Experimental results demonstrate that our DDVT outperforms state-of-the-art methods on several widely-used benchmarks. 

\end{abstract}
\section{Introduction}
\label{sec:1}
 Answer grounding in Visual Question Answering (VQA) represents a vital intersection between computer vision and natural language processing~\cite{101,102,103}. This multimodal task aims to not only provide accurate answers to questions based on visual data but also identify or "ground" the regions in the image that support the answer, which involves the extraction of feature data from the vision using image processing techniques. The primary challenge lies in the system's ability to understand the nuanced interplay between visual and textual elements, enabling it to reason and make inferences based on both.
It holds immense potential for diverse real-life applications, including answering questions from visually impaired individuals~\cite{7}, aiding radiologists in early diagnosis of life-threatening diseases~\cite{9}, and enhancing human-computer interaction~\cite{11}.
As VQA systems continue to advance, the need for interpretability in various applications becomes increasingly evident. 
 An ideal VQA system for such purposes should not only provide accurate answers but also offer a mechanism for answer verification.This is necessary to ensure that the answers are derived based on the correct image location, rather than language biases or other factors.

Efforts have been made recently for this task. 
MAC-Caps~\cite{4} aims to enhance answer localization accuracy by generating visual attention maps alongside textual answers in evaluation systems. 
Similarly, approaches like LXMERT~\cite{Lxmert} and SDCAM~\cite{SDCAM} produce textual answers while simultaneously identifying the corresponding grounded answer regions in images. 
While these methods can generate attention maps or bounding boxes somewhat related to the question, they exhibit certain limitations.
Firstly, existing methods may not effectively capture the intricate relationship between textual features and visual ones.
Secondly, these methods might overlook the crucial task of aggregating pixel-level and region-level features.

To overcome these limitations, we propose a Dynamic Dual-level Vision Transformer Fusion Network (DDVT) that jointly predicts textual answers and generates accurate grounding masks. Through a novel question-guided region-level selection strategy and a cross-modal multi-scale aggregation mechanism, our method adaptively fuses pixel-level and region-level visual information, effectively localizing regions that correspond to textual answers.

\begin{itemize}
\item 
We propose a dual-level visual transformer framework for answer grounding, which constructs direct flows from pixel-level features to region-level features at multiple levels, thus facilitating complementary information aggregation from multi-level features.
\item We propose a question-guided dynamic regional-level module to effectively locate region-level objects based on questions and dynamically select masks of different resolutions. 

\item We propose a cross-modal multi-scale fusion module to adaptively aggregate pixel-level information and region-level content guided by language in the image, allowing for the interaction and fusion of multimodal information from different levels.
\end{itemize}

\section{Related work}\label{sec:Related work}
\label{sec:2}
\subsection{Visual Question Answering and Answer Grounding}

In recent years, methods for the VQA task have relied on object-level features as input to enhance the accuracy of the system\cite{4,32,33,34,35,36}. Those features are extracted from pre-trained object detectors. This makes the VQA task easier and usually performs better than spatial or appearance features, but it also adds an additional preprocessing step (detecting objects) to the pipeline. Furthermore, since pre-training relies on object classes in the training dataset, these feature mappings are pre-trained with relevant object categories. This limits the system's learning scope to known object classes or requires annotated regions for related objects and pre-training an object detector for them. All these methods usually focus only on the accuracy of the textual answer output or generate visual attention maps to indicate whether the answer is based on correct visual evidence, but they without providing evaluate the grounding of the answers.

Recently, researchers have proposed that a good VQA model should provide correct answers by attending to the correct visual evidence\cite{37}. Therefore, it is crucial to ensure that the model's visual attention is correctly focused when generating answers. To address this issue, several datasets have been introduced that provide grounding labels, which locate each visual object relevant to the language question and answer. Some of these datasets were created by tracking people's attention when observing visual questions or by collecting bounding boxes of relevant visual answers. Additionally, datasets like VizWizground~\cite{grounding} and VQS~\cite{6} provide mask annotations for visual answers that are related to the questions. Notably, VizWizground is the first answer grounding dataset that reflects real-world VQA scenarios. By using this dataset, we can evaluate not only the accuracy of the answers but also assess the accuracy of the grounding answers using the Intersection over Union (IOU) metric.

\begin{figure*}[!t]
  \centering
   \includegraphics[width=0.95\linewidth]{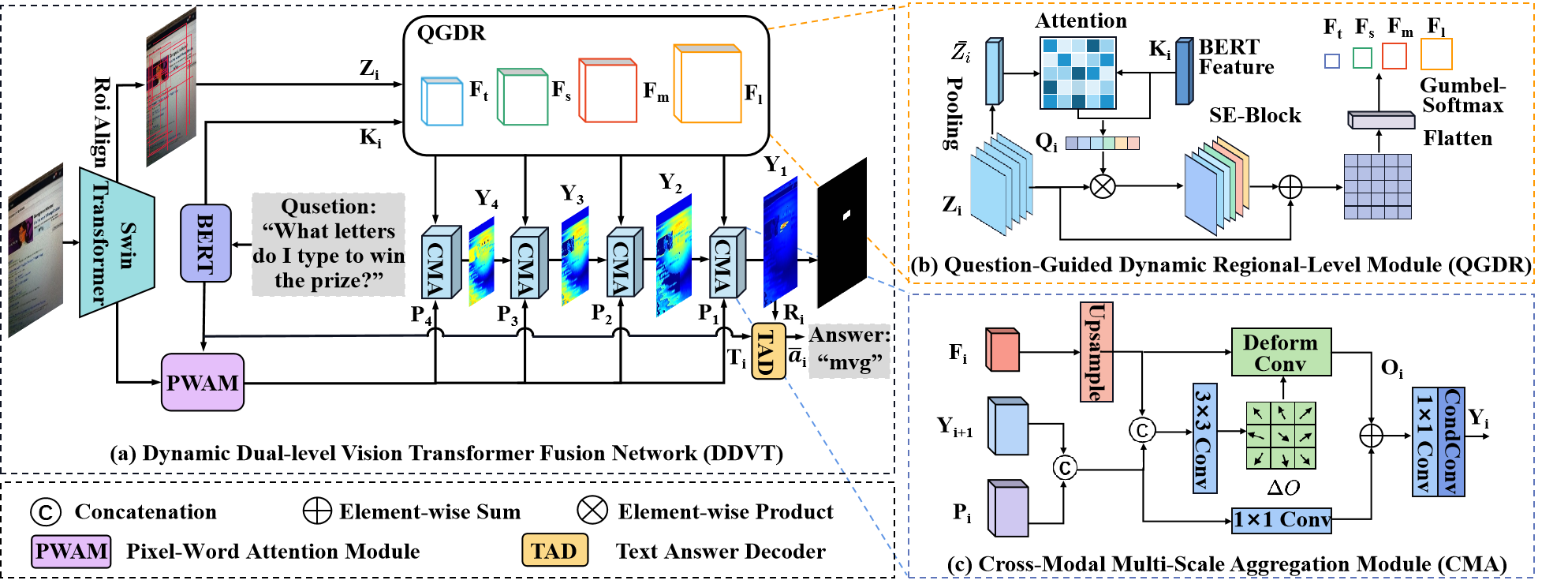}
    \vspace{-4mm}
   \caption{
   The overall architecture of Dynamic Dual-level Vision Transformer Fusion Network (DDVT).
   we leverage the Swin Transformer to extract visual features and the BERT model to obtain language question features. 
   These are then fused using our proposed Question-Guided Dynamic Regional-Level (QGDR) module, enabling dynamic object region identification and mask selection based on questions.
   Additionally, we introduce a Cross-Modal Multi-Scale Aggregation (CMA) module to adaptively merge pixel- and region-level information, guided by the language. 
   Ultimately, DDVT predicts the answer grounding mask and generates the corresponding answer based on the visual image and question input.}
   \label{fig:The overall architecture of Dynamic Dual-level Vision Transformer Fusion Network}
   \vspace{-5mm}
\end{figure*}
\subsection{Dynamic Networks}
Dynamic networks are able to adaptively adjust the network structure based on input features. Although traditional static networks can maintain inference efficiency through classification selection or budget constraints, the network parameters and computation are very redundant during training, which leads to a huge demand for experimental resources. Unlike traditional static network structures, dynamic networks can dynamically select or adjust the structure or parameters of their instances during the inference process to obtain better representability, adaptability, compatibility, and generality \cite{25,42,43,44,45}, which are more compatible with various situations during practical applications, so they have been widely studied and applied in recent years.

Dynamic networks can be classified into three categories according to their structural design: sample adaptive, spatial adaptive, and temporal adaptive \cite{28,29}. The dynamic network methods commonly used in the VQA field are generally sample adaptive, such as the dynamic routing method used by TARA~\cite{26} network proposed by Zhao et al. In the field of image segmentation, spatial adaptive methods have been proven effective for object localization and segmentation. For instance, Huang et al.~\cite{skipnet} propose a multi-scale dense network with multiple classifiers for allocating different computations for “easy” and “hard” samples. The DRNET~\cite{27} network proposed by Li et al. to reduce redundancy in the input resolution of modern CNN networks. The DynaMask~\cite{2} network proposed by Li et al. utilizes dynamic masks of different resolutions to segment different instances. We are inspired to improve the object localization and grounding answer segmentation performance of VQA systems without increasing the computational load. This motivates our study where we adopt a spatial adaptive network, the first attempt to design a question-guided dynamic region-level feature module (Section 3.1). Based on question guidance, it performs object localization and adaptively assigns masks of suitable resolutions for different target segmentation difficulties.

\section{Method}\label{sec:Method}
\label{sec:3}
Our goal aims to local the answer grounding in one image in the visual question answering system.
To that end, we propose a Dynamic Dual-level Vision Transformer Fusion Network (DDVT).
Fig.~\ref{fig:The overall architecture of Dynamic Dual-level Vision Transformer Fusion Network} (a) shows the overall architecture of our DDVT.
We first exploit the Swin Transformer~\cite{17} as the baseline backbone to extract the visual features that are further utilized to generate Roi Align features, and utilize the BERT model~\cite{15} to generate language question features.
Then we adopted the Pixel-Word Attention Module (PWAM)~\cite{100} , which utilizes attention operations between pixel-level visual features and linguistic cues to ensure spatial alignment between visual image areas and corresponding textual descriptions. Specifically, this module builds a cross-modal interaction map by computing attention scores between visual feature maps and word embeddings, selectively enhancing the relevant visual features based on linguistic context, which effectively bridges semantic gaps between the modalities.
To better fuse the language question features and visual ones, we propose a Question-Guided Dynamic Regional-Level module (QGDR), which is introduced in Sec.~\ref{sec:Question-Guided Dynamic Regional-Level Module}, to effectively locate region-level objects based on questions and dynamically select masks of different resolutions.
To further improve the interaction of multi-modal information from different levels, we propose a Cross-Modal Multi-Scale Aggregation module (CMA), which is introduced in Sec.~\ref{sec:Cross-Modal Multi-Scale Aggregation Module}, to adaptively aggregate pixel-level information and region-level content from the outputs of PWAM~\cite{100} and QGDR guided by language in the image.

Finally, our DDVT predicts the answer grounding mask and generates the answer from the give the visual image and question.

\subsection{Question-Guided Dynamic Regional-Level Module}\label{sec:Question-Guided Dynamic Regional-Level Module}

We propose a Question-Guided Dynamic Regional-level (QGDR) module to effectively locate region-level objects based on questions to improve the quality of grounded answer segmentation, as shown in Fig.~\ref{fig:The overall architecture of Dynamic Dual-level Vision Transformer Fusion Network} (b).
First, the ROI-aligned regional feature $Z_{i}$ extracted from the Swin Transformer~\cite{17} is average pooling to obtain $\bar{Z_{i}} $, which is then combined with the question feature $K_{i}$ extracted from the BERT~\cite{15}. 
Then, the $\bar{Z_{i}}$ and $K_{i} $ are performed within cross-modal attention:
\begin{equation}
  Q_{i}= \text{softmax}\left(\frac{\bar{Z} _{i}\bar{K}_{i} ^{T}  }{\sqrt{D_{i} }}  \right)\hat{K_{i}} ,      
\label{eq:important}
\end{equation}
where $Q_{i} \in \mathbb{R}^{C\times d_{v}}$ and $D_{i}$ means the number of dimension of the head. 
Then, the obtained is performed with a global pooling operation on $Q_{i}$ to get the feature weight $\bar{Q_{i}} \in \mathbb{R}^{C\times 1\times 1} $. 
The $\bar{Q_{i}} $ is fed into the channel attention module SE-block~\cite{21} to weigh the different channels of visual information. 
Finally, several convolutional and fully-connected layers are used to classify to obtain different sizes of $F_{i}$.

The QGDR module is essentially a lightweight classifier, aiming to accurately locate and segment the best mask resolution from the $k$ candidate targets of different scales in a cost-effective manner. 
QGDR divides $F_{i}$ into 4 categories of region-level hierarchy feature $\left\{F_{t}, F_{s}, F_{m}, F_{l}  \right \} $, with spatial resolutions doubling from $F_{t}$ to $F_{l}$. 
%

It then performs softmax operation to output a probability vector $\varepsilon _{k}=\left [ \varepsilon _{1} ,\cdots,\varepsilon _{k}\right]$. 
 Each element of this vector represents the probability of selecting the corresponding candidate resolution. The soft output $\varepsilon ^{k} $ of QGDR should be transformed into a one-hot prediction, denoted as $H=\left [ h _{1}, \cdots, h _{k} \right ]$, where $ h _{i}\in \left \{ 0, 1 \right \} $. 
 This process can be achieved through discrete sampling, and in this paper, Gumbel-Softmax~\cite{22} is utilized for gradient-based back propagation to update QGDR. The specific formula is as follows:
\begin{equation}
     h_{i} =\frac{exp \left( \left(log \varepsilon _{k} +g_{i}  \right)   /\tau \right) }{\sum_{k'}^{} exp \left (\left (log \varepsilon _{k}  +g_{i} \right)    /\tau \right)} , 
      \label{eq:also-important}
 \end{equation}
 where $\tau $ denotes a temperature parameter; 
  When $\tau$ approaches 0, the Gumbel-softmax is close to one-hot. 
 $g_{i}$ are samples drawn from the Gumbel $(0, 1)$ distribution.

 \subsection{Cross-Modal Multi-Scale Aggregation Module}\label{sec:Cross-Modal Multi-Scale Aggregation Module}
 
To better align the cross-modal features at multiple scales, we introduce a cross-modal multi-scale aggregation module (CMA), as shown in Fig.~\ref{fig:The overall architecture of Dynamic Dual-level Vision Transformer Fusion Network} (c).
Firstly, we use a deconvolution layer to enlarge the spatial size of $F_i$ which is the output of QGDR. 
Then, $F_{i}$ is concatenated with $P_{i}$ which is generated by the PWAM~\cite{100} to explore the pixel-level interaction between visual and language features.
Next, we use 3$\times$3 convolution on the concatenated features to obtain the offset mapping. 
After that, the deconvolution feature of $F_{i}$ is aligned with the concatenated feature via a Deformable convolution~\cite{deformable}. 
Then, the aligned feature is added to $P_{i}$, which would be passed by a $1\times1$ convolution and CondConv~\cite{condconv} which serves to pay more attention to the salient parts of the object. 

\subsection{Loss Funcion}\label{ssec:Objective Function }
We use Mask Loss $\mathcal{L} _{mask}$, Edge Loss $\mathcal{L} _{edge}$, Budget Constraint $\mathcal{L} _{budget}$, Text Answer Loss $\mathcal{L}_{text}$ to train our model within an end-to-end fashion:
\begin{equation}
\mathcal{L}_{total} =\mathcal{L}_{mask}+  \lambda_{1} \mathcal{L}_{edge}+ \lambda_{2} \mathcal{L}_{budget} + \mathcal{L}_{text},
    \label{eq:also-important}
\end{equation}
where $\lambda _{1} $ and $\lambda _{2} $ are the trade-off hyper-parameters.

\noindent \textbf{Mask Loss.}
Given a VQA instance, we first predict its mask switching state $H= \left [ h_{1}, \dots, h_{k} \right ]$  by QGDR, and obtain a group of mask prediction maps at $k$ different resolutions $\left \{ m_{i}^{1},\dots, m_{i}^{k}  \right \} $ by passing this instance through different stages of decoding. The mask loss function is defined as:
\begin{equation}
   \mathcal{L} _{mask} =\sum_{i=1}^{N} \sum_{k=1}^{K} h_{i} \mathcal{C}   \left ( \hat{m}_{i}^{k}  ,m _{i}   \right ),
    \label{eq:also-important}
\end{equation}
where $\hat{m}_{i}^{k}$ denotes the $k$-th mask prediction ground truth answer , and ${m} _{i}$ represents corresponding ground truth mask grid. 
$h_{i}$ is the indicator for whether the $k$-th mask resolution is selected as the output resolution. 
$\mathcal{C} $ is the binary cross-entropy loss.

\noindent \textbf{Edge Loss.} The edge loss for the edge mapping at different resolutions is defined as follows:

\begin{equation}
    \mathcal{L} _{edge} =\sum_{i=1}^{N} \sum_{k=1}^{K} h_{i} \mathcal{C}   \left (  \hat{e}_{i}^{k}   ,{e} _{i}   \right ),
     \label{eq:also-important}
\end{equation}
where ${e}_{i}$ denotes the ground-truth grounding answer edge, which is generated by first applying the Laplacian operator on the ground-truth grounding answer mask ${m}_{i}$ to obtain a soft edge map and then converting it into a binary edge map by thresholding.

\noindent \textbf{Budget Constraint.}
We adopt budget constraints~\cite{2} to train QGDR. Specifically, let $C$ represent the corresponding computational cost for the selected mask resolution. A penalty is added to the model when the expected deviation $\mathbb{E\left (\mathrm {C} \right)}$ from the current batch data exceeds the target deviation (represented as $\mathrm {C} _{t} $) to control the computational cost:
\begin{equation}
    \mathcal{L} _{budget} =max\left(\frac{\mathbb{E\left ( \mathrm {C} \right) }}{\mathrm {C}_{t} }-1,0 \right ),
    \label{eq:also-important}
\end{equation}
\noindent \textbf{Text Answer Loss.}
Textual answers are constrained by binary cross-entropy loss and obtained through the text answer decoder~\cite{Lxmert}(TAD):
\begin{equation}
\mathcal{L}_{text}  = \mathcal{C}\left(\sigma(\bar{a}_{i}), {a} _{i}\right).
\end{equation}
where $\bar{a}_{i}$ denotes the textual answers prediction; $a_{i}$ represents corresponding truth textual answers;
$\sigma$ means the sigmoid function.


\begin{table}[!t]
  \centering
\caption{Comparison results with the state-of-the-art methods on the VizWizGround~\cite{grounding} test set and VQS~\cite{6} val set. The pre-trained represents whether the model adopts a large-scale dataset for pre-training or not.}

\begin{tabular}{lllccccll}
\toprule
\multicolumn{3}{l}{\multirow{2}{*}{Method}} & \multicolumn{1}{c}{\multirow{2}{*}{Pre-trained}} & \multicolumn{4}{c}{IOU}                 \\ \cline{5-8} 
\multicolumn{3}{c}{}       & \multicolumn{1}{c}{}                             & VizWizGround  & \multicolumn{3}{c}{VQS} \\ 
\toprule
\multicolumn{3}{l}{LXMTRT~\cite{Lxmert}}  & \CheckmarkBold & 22.09 & \multicolumn{3}{c}{-}     \\  
\multicolumn{3}{l}{Mac-Caps~\cite{4}} & \XSolidBrush & 27.3  & \multicolumn{3}{c}{-}     \\
\multicolumn{3}{l}{UNIFIED~\cite{unifiedio}}  &    \CheckmarkBold         & 54.7  & \multicolumn{3}{c}{-}     \\
\multicolumn{3}{l}{DDTN~\cite{40}}     &    \XSolidBrush                  & 53.4  & \multicolumn{3}{c}{-}     \\
\multicolumn{3}{l}{DFAF~\cite{DFAF}}     &   \XSolidBrush                   & -     & \multicolumn{3}{c}{17.5}  \\
\multicolumn{3}{l}{MCAN~\cite{MCAN}}     &   \XSolidBrush                   & -     & \multicolumn{3}{c}{23.91} \\
\multicolumn{3}{l}{BUTD~\cite{Butd}}     &    \XSolidBrush                  & -     & \multicolumn{3}{c}{33.97} \\
\multicolumn{3}{l}{SDCAM~\cite{SDCAM}}    &    \XSolidBrush                  & -     & \multicolumn{3}{c}{37.93} \\
\toprule
\multicolumn{3}{l}{\textbf{DDVT (Ours)}}     &      \XSolidBrush   & \textbf{65.3}  & \multicolumn{3}{c}{\textbf{43.47}} \\ 
\toprule
\end{tabular}
\label{tab:Comparison results with the state-of-the-art methods on the VizWizGround}
\vspace{-2mm}
\end{table}

\begin{figure*}[t]
  \centering
    \includegraphics[width=0.9\linewidth]{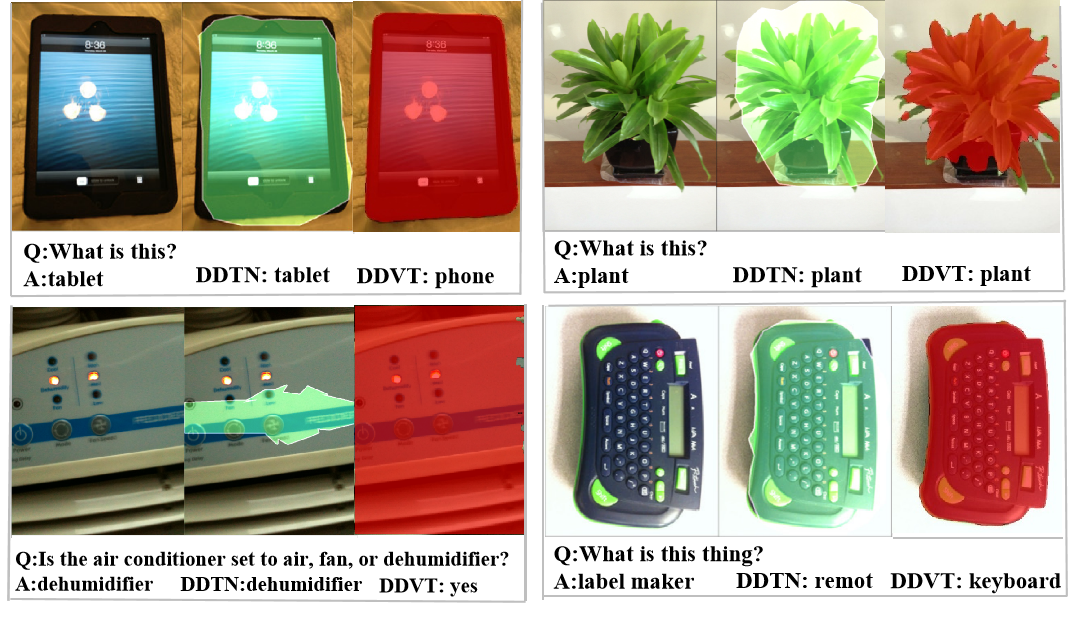}
     
   \caption{Predicted masks generated by DDTN~\cite{40} and our DDVT on the VizwizGround~\cite{grounding} val set.}
   \label{fig:Visualized results of the predicted 
 masks generated by DDVT and DDTN}

\end{figure*}

\section{Experiments}\label{sec:Experiments}

\subsection{Datasets and metrics}
We conduct experiments on the VizwizGround~\cite{grounding} and VQS~\cite{6} datasets. The VizWizGround dataset consists of 6494 images for training, 1131 images for validation, and 2373 images for online testing. Each example includes a question-answer pair and the corresponding instance mask. This dataset comprises 9998 VQA triplets and 9998 Grounding pairs. Compared to other VQA datasets, it has lower image quality, making the task more challenging. The VQS dataset contains 37868 images, 96508 questions, 108537 instance segmentations, and 43725 bounding boxes. The accuracy of grounding answers based on textual responses is used to evaluate the inference performance of models [35].

For grounding answer evaluation, the IoU is used to measure the similarity between each binary segmentation mask and the ground truth grounding segmentation answer. The average IoU score is calculated for all test examples. The range of the Q value is from 0 to 1, with higher values indicating better performance. For the validation set, we also evaluate using the general metrics for detection and localization tasks: mAP@IoU based on the COCO evaluation protocol, and the average AP values with IoU thresholds ranging from 0.5 to 0.95 with a step size of 0.05.
\subsection{Implementation Details}\label{ssec:subhead}
We implement the method by PyTorch~\cite{13}.
%
%
Swin Transformer~\cite{17} is adopted as the backbone, which is initialized with classification weights on ImageNet22K~\cite{16}. 
The language encoder is based on the BERT model~\cite{15} initialized with official pre-trained weights. 

We use Adam~\cite{19} optimizer with an initial learning rate of $0.00005$ for training.
The model training is performed on $4$ GPUs for $100$ epochs with a batch size of $16$. 
The $K_i$ is set to $512$.
The $\lambda_{1}$ and $\lambda_{2} $ are set to $0.1$ and $0.4$, respectively.

\subsection{Main Results}\label{ssec:subhead}

We conduct experiments using the 
VizWizGround dataset~\cite{grounding}
and the VQS dataset~\cite{6}. 
Since these datasets exclusively support IOU 
comparisons of grounding answers and do not offer text answer verification, our method is 

compared with existing approaches based on IOU metrics. 
The results in Tab.~\ref{tab:Comparison results with the state-of-the-art methods on the VizWizGround} demonstrate that our method achieves a grounded answer accuracy of $65.3\%$ on the VizWizGround test set~\cite{grounding}. 
This represents a significant $12\%$ improvement in accuracy compared to the latest method, DDTN~\cite{40}. 
Notably, our method achieves state-of-the-art performance even without extensive pre-training, surpassing the Unified model by $11\%$.
On the VQS~\cite{6} dataset, our method attains an accuracy of $43.47\%$, a substantial improvement over previous methods, highlighting the effectiveness of our approach.

To provide further evidence effectiveness of our method, we visualize the estimated masks and predicted text answers in Fig.~\ref{fig:Visualized results of the predicted masks generated by DDVT and DDTN}. It is evident that our model produces more detailed object edge segmentation. 
For instance, in the second sample of the first row, our method demonstrates superior ability in modeling complex object contours, as seen in the ``plants" segmentation. 
Additionally, in the last example, while DDTN~\cite{DLVT} struggles to accurately locate the object, our model excels in capturing global information. 
However, it is worth noting that our method still faces challenges in generating highly accurate textual answers, particularly when distinguishing between objects with similar appearances, such as ``label maker" and ``keyboard".

Moreover, to address the discrepancy between our model’s outstanding localization performance and relatively lower textual answer accuracy presented in Fig.~\ref{fig:Visualized results of the predicted masks generated by DDVT and DDTN}, we perform an additional analysis. This difference primarily arises because the DDVT model emphasizes more on effectively localizing complex object boundaries and adopts a robust fusion scheme between pixel-level and region-level features guided explicitly by linguistic cues. Such a sophisticated localization mechanism might overspecialize on grounding answers and cause suboptimal final textual predictions, especially in cases with visually or semantically ambiguous objects. Future work will explore balancing grounding and classification tasks to minimize this discrepancy.

\begin{figure}[!t]
\centering
\includegraphics[width=\linewidth]{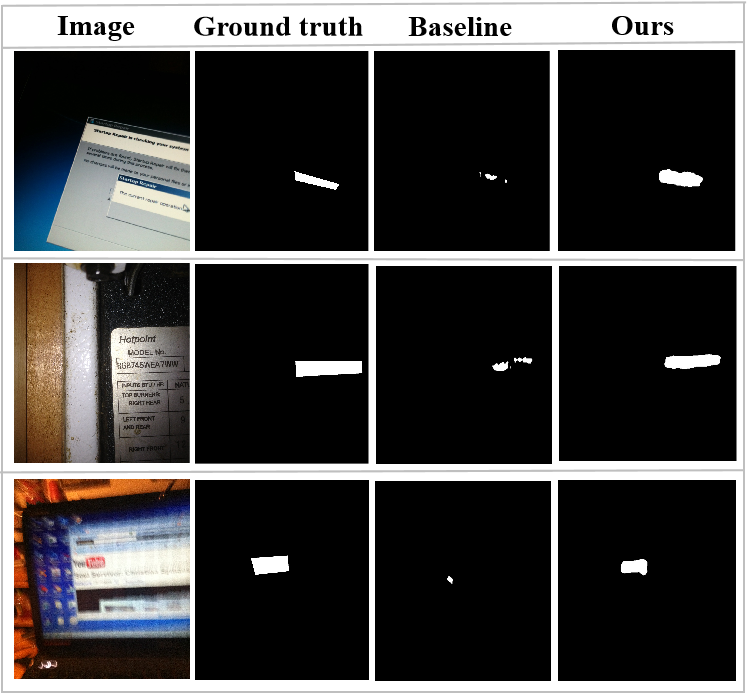}
\caption{Visualized predictions from the VizWizground~\cite{grounding} val set. Baseline is LAVT~\cite{100}.}
\label{fig:Visualized predictions on two examples from the VizWizground}

\end{figure}

\subsection{Ablation Study}
\label{ssec:subhead}
To validate the effectiveness of our question-guided dynamic regional-level module (QGDR) and cross-modal aggregation module (CMA), we conducted ablation experiments. 
Tab.~\ref{tab:Ablation results on the VizWizGround} clearly demonstrates their positive impact on performance.
Disabling the QGDR resulted in a notable decrease in IoU by an absolute value of $4.17$ (see Tab.~\ref{tab:Ablation results on the VizWizGround} (a) vs. (d)). 
Furthermore, when we replace the CMA with a simple concatenation between $F_i$ and $P_i$, we observe a decrease in IoU by $3.57$ (see Tab.~\ref{tab:Ablation results on the VizWizGround} (c) vs. (d)). 
This change also leads to a drop in accuracy by $3$ to $4$ points across the two IoU thresholds. 
These results strongly indicate the significant performance benefits benefits from the proposed QGDR and CMA.
Additionally, Tab.~\ref{tab:Ablation results on the VizWizGround} highlights the utility of the used PWAM in performance enhancement (see Tab.~\ref{tab:Ablation results on the VizWizGround} (b) vs. (d)).

To provide further insight into our method's effectiveness, we visually present segmentation results using ground truth answers for experiments involving small targets in Fig.~\ref{fig:Visualized predictions on two examples from the VizWizground}. 
In these cases, the baseline struggles to accurately segment any objects. In contrast, our method closely approximates the ground truth answers, demonstrating its robustness and proficiency in segmenting small target answers.

\begin{figure*}
  \centering
    \includegraphics[width=1\linewidth]{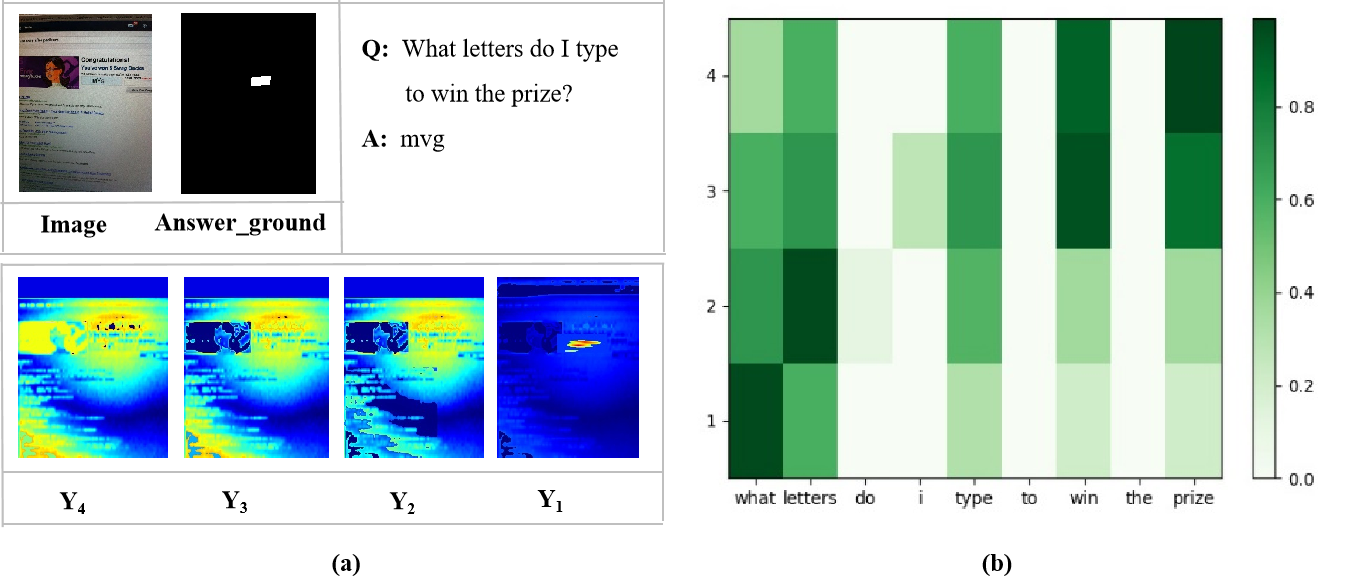}
   \caption{Visualized predictions and feature maps on an example from the VizWizgroun validation set. (a) shows from left to right, the left-most column illustrates the input image and the truth answer ground mask.  We visualize the predicted mask and the feature maps used for final classification ($i.e.$, $Y_4$,$ Y_3$, $Y_2$, and $Y_1$) from left to right. (b) shows the output of attention to questions in word pixel attention, with the vertical axis corresponding to each of the four aggregation stages. }
   \label{fig8}
\end{figure*}

\begin{figure}
  \centering
    \includegraphics[width=0.8\linewidth]{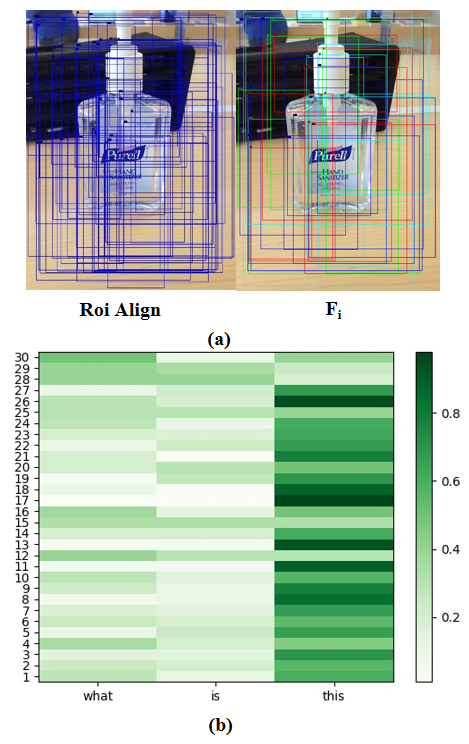}
    \vspace{-1.5mm}
   \caption{(a) demonstrates the output bounding boxes of 50 after being processed through ROI Align, as well as the output bounding boxes of 30 after being scored by QGDR. (b) displays the cross-attention maps generated by QGDR. Each index from [1-30] corresponds to each object in the subgraph (a).}
   \label{fig3}
\end{figure}

\begin{table}[!t]
\caption{Ablation results on the VizWizGround~\cite{grounding} val set.}
\begin{center}
\resizebox{0.7\linewidth}{!}{
\begin{tabular}{l|ccc|cccc}
\toprule
ID & QGDR     & PWAM     & CMA      & IOU   & mAP50 & mAP   \\ 
\toprule
(a) &\XSolidBrush       & \CheckmarkBold & \CheckmarkBold & 69.44 & 74.09 & 60.28 \\
(b)& \CheckmarkBold &      \XSolidBrush    & \CheckmarkBold & 65.35 & 70.56 & 49.67 \\
(c) &\CheckmarkBold& \CheckmarkBold &      \XSolidBrush    & 70.04 & 75.86 & 60.24 \\
\toprule
\textbf{(d)} &\CheckmarkBold & \CheckmarkBold & \CheckmarkBold & \textbf{73.61} & \textbf{79.84} & \textbf{65.16} \\ 
\toprule
\end{tabular}
}
\end{center}
\label{tab:Ablation results on the VizWizGround}

\end{table}

\begin{table}[!t]
\caption{Performance of using different grounding answer mask sizes.}
\begin{center}
\resizebox{0.7\linewidth}{!}{
\begin{tabular}{cccccc}
\toprule
Mask Size & FLOPs(G) & IOU   & mAP   & mAP50 & mAP75 \\ \hline
14*14     & 13.02  & 70.82 & 62.19 & 77.15 & 60.17 \\
28*28     & 14.67  & 73.58 & 65.02 & 79.73 & 63.78 \\
56*56     & 16.56  & 73.71 & 65.18 & 79.86 & 63.91 \\
112*112   & 19.10  & 74.68 & 67.01 & 81.10 & 66.14 \\ \hline
\toprule
\end{tabular}
}
\end{center}
\label{tab:t11}
\vspace{-2mm}
\end{table}

\subsection{Discussion}\label{ssec:discuss}
 
 {\bf The impact of the mask size in the dynamic multi-scale module.}  We further investigate how the mask size impacts both model accuracy and computational complexity. An effective model should strike a balance between performance and computational demands. 
 To this end, we conduct several experiments by fixing different mask sizes.
 QGDR module outputs four different candidate mask resolutions $\left [ 14\times 14, 28\times 28, 56\times 56, 112\times112 \right ]$. 
 We choose one of them as the uniform output mask size and present the corresponding results in Tab.~\ref{tab:t11}. It is evident that models with larger mask resolutions can achieve higher segmentation performance, but this also significantly increases computational costs. Furthermore, as the mask size further increases, the performance tends to saturate. Therefore, dynamic methods can achieve comparable performance to the baseline at a lower cost.

{\bf Visualized predictions.} To further validate the effectiveness of our proposed method of location grounding answer information via a dual-level vision Transformer fusion Network. In Fig.~\ref{fig8} (a) shows the visual attention maps for each stage after the multi-scale aggregation module, specifically for the Y4, Y2, Y3, and Y1 stages. Combining this with the word-pixel attention in Fig.~\ref{fig8} (b), we can observe that the visual attention at each stage progressively aligns with the attention output from the question. Ultimately, it accurately localizes the correct grounding answer. From Fig.~\ref{fig8} (a), we can observe that the higher-level feature maps ($i.e.$,  $\left\{Y_{4}, Y_{3}, Y_{2}, \right \} $) in our full model can accurately locate the semantic concept given in question, while the low-level feature maps ($i.e.$, $\left\{Y_{1}, \right \} $) contain rich boundary information important to binary segmentation.

{\bf Visualisation QGDR.} In order to further validate the effectiveness of our proposed QGDR for localizing region-level object information. Visualizes the input and output of the QGDR module, as well as the intermediate processing in Fig.~\ref{fig3}. The visualization in Fig.~\ref{fig3} (a) displays 50 boxes detected by the Roi Align part, while Fig.~\ref{fig3} (a) shows the output after passing through the QGDR module. The 30 detected boxes are labeled with numbers, and the four different colors represent the four different resolution sizes of the output. Fig.~\ref{fig3} (b) presents the output of the cross-attention part, demonstrating the scores assigned to the 30 boxes based on different words. We can observe that the word "this" is a crucial localizing word, so objects with higher scores in this region are more likely to be the correct grounding answers. These results indicate that QGDR is capable of accurately localizing objects relevant to the answer.



\section{Conclusion}
\label{sec:page}
We have proposed a Dynamic Dual-level Vision Transformer Fusion Network (DDVT) for grounded visual question answering. 
Our framework leverages multi-level visual transformers to collectively encode cross-modal multi-scale inputs. 
It employs language guidance to dynamically pinpoint visual information across multiple scales and aggregates cross-modal information effectively.
Experimental results on two widely used benchmark datasets have demonstrated the superiority of our DDVT over existing techniques.\\

\noindent 
\textbf{Acknowledgments} This work was supported in part by National Natural Science oundation of China (Grant No. 62502064),
‌Liaoning Province Science and Technology Joint Plan  (Grant No. 2025JH2/101800417),
Educational Department of Liaoning Province, China (Grant No. LJ222511258003),
Liaoning Provincial Key Research and Development Joint Program (Grant No. 2025110219-JH2/1018), 
Interdisciplinary project of Dalian University (Grant No. DLUXK-2025-QN-020),
111 Center (Grant No. D23006).
%
%
%
\bibliographystyle{splncs04}
%





\bibliography{refe.bib} 

\end{document}